\documentclass[10pt,journal,compsoc]{IEEEtran}

\ifCLASSOPTIONcompsoc
\usepackage[nocompress]{cite}
\else
\usepackage{cite}
\fi

\ifCLASSINFOpdf
\usepackage[pdftex]{graphicx}
\else
\usepackage[dvips]{graphicx}
\fi

\usepackage{amsmath}
\usepackage{amssymb}
\usepackage{booktabs}
\usepackage{multirow}
\usepackage{color}

\begin{document}

\title{Linguistically Driven Graph Capsule Network for Visual Question Reasoning}

\author{Qingxing Cao, Xiaodan Liang, Keze Wang, and Liang Lin
\IEEEcompsocitemizethanks{\IEEEcompsocthanksitem
Q. Cao, X. Liang, and L. Lin are with Sun Yat-sen University, Guangzhou, China. Email:  caoqx@mail2.sysu.edu.cn; xdliang328@gmail.com; linliang@ieee.org. \protect
\IEEEcompsocthanksitem K. Wang is with Department of Statistics, University of California, Los Angles, U.S.\protect\\
Email:   kezewang@gmail.com.}}
\markboth{IEEE Transactions on Pattern Analysis and Machine Intelligence,~Vol.~X, No.~X, XXX}%
{Shell \MakeLowercase{\textit{et al.}}: Bare Advanced Demo of IEEEtran.cls for IEEE Computer Society Journals}

\IEEEtitleabstractindextext{%
\begin{abstract}
Recently, studies of visual question answering have explored various architectures of end-to-end networks and achieved promising results on both natural and synthetic datasets, which require explicitly compositional reasoning. However, it has been argued that these black-box approaches lack interpretability of results, and thus cannot perform well on generalization tasks due to overfitting the dataset bias. In this work, we aim to combine the benefits of both sides and overcome their limitations to achieve an end-to-end interpretable structural reasoning for general images without the requirement of layout annotations. Inspired by the property of a capsule network that can carve a tree structure inside a regular convolutional neural network (CNN), we propose a hierarchical compositional reasoning model called the ``Linguistically driven Graph Capsule Network'', where the compositional process is guided by the linguistic parse tree. Specifically, we bind each capsule in the lowest layer to bridge the linguistic embedding of a single word in the original question with visual evidence and then route them to the same capsule if they are siblings in the parse tree. This compositional process is achieved by performing inference on a linguistically driven conditional random field (CRF) and is performed across multiple graph capsule layers, which results in a compositional reasoning process inside a CNN. Experiments on the CLEVR dataset, CLEVR compositional generation test, and FigureQA dataset demonstrate the effectiveness and composition generalization ability of our end-to-end model.
\end{abstract}

}

\maketitle

\IEEEdisplaynontitleabstractindextext

%
\IEEEpeerreviewmaketitle

\ifCLASSOPTIONcompsoc
\IEEEraisesectionheading{\section{Introduction}\label{sec:introduction}}
\else
\section{Introduction}
\label{sec:introduction}
\fi

%
%
%
%


\IEEEPARstart{V}{isual} question answering (VQA) that predicts the correct answer given an image and a textual question requires the agent to possess the co-reasoning capability to connect vision and linguistic knowledge. A desirable agent must be capable of perceiving both visual evidence and language semantic meaning and then performing logical inference based on its observations.
This task combines computer vision and natural language processing and requires both the perception and inference abilities, causing it to attract increased interest in recent years.

\begin{figure}[ht!]
	\centering
	\includegraphics[width=1.00\columnwidth]{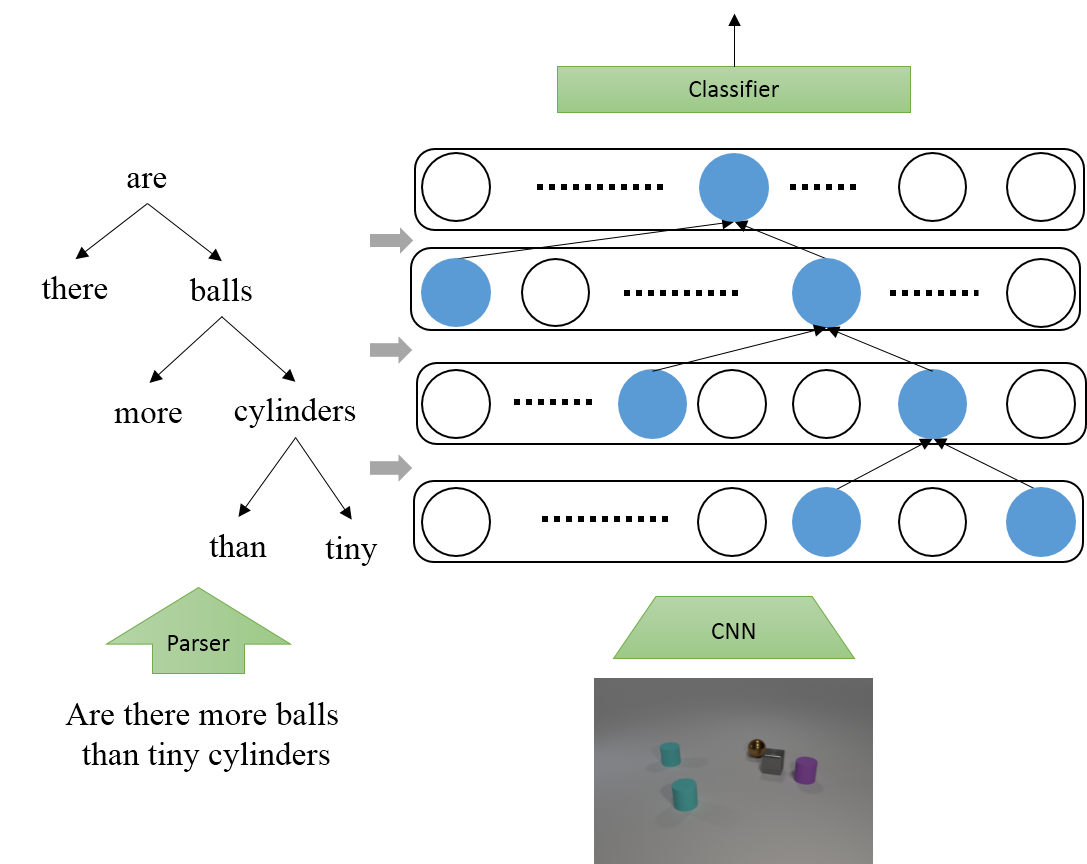}
	\caption{Our proposed Linguistically driven Graph Capsule Network (LG-Capsule) aims to carve a tree structure inside a CNN by merging capsules from the bottom to the top following the guidance of a linguistic parse tree. As shown, each row represents a capsule layer and each circle represents a capsule. This compositional process can be performed across multiple graph capsule layers and results in the tree structure indicated by the blue circle.    
	}
	\label{fig:intro}
\end{figure}

The pioneering studies on VQA~\cite{Seq2Seq, HiCoAtt, MLB} focused on utilizing neural networks and learning the mapping from inputs to answer outputs. Despite the steadily improving performance of current pipelines, they have exhibited fundamental limitations, whereby an end-to-end neural network often learns the dataset bias instead of fully understanding the image and language~\cite{balanced_vqa_v2, ECCV16baseline}. For instance, even on a dataset~\cite{clevr} with minimal biases and intricate compositional questions, the end-to-end neural networks~\cite{FiLM} with a simple fusion architecture still achieve exceptionally good results without using additional layout annotations.
In addition to the end-to-end neural networks, numerous efforts have been devoted to structural modeling~\cite{GraphVQA, SAVQA} to pursue the interpretable reasoning capability. A series of studies on modular networks~\cite{e2emn,inferring2017} were proposed to first detect atomic elements and then integrate elements to infer the final results. However, such studies rely heavily on the ground truth layouts, and their performance drops rapidly if the layout annotations are inaccessible. Such drawbacks severely hinder the application potential of these studies on general and unconstrained VQA tasks for natural image scenarios.

As a matter of fact, humans can recognize novel concepts by incorporating learned concepts~\cite{lake2015human}. This compositional generalization ability allows people to solve a plethora of problems using a limited set of basic skills and is one of the major differences between human intelligence and the current deep neural networks. To address the composition issue, there are successful hierarchical and compositional models~\cite{DPM} that explicitly represent an entity with parts. 
Meanwhile, the capsule network~\cite{capsule, EMcapsule} models different parts and properties of an entity by grouping the feature channels into multiple capsules. It advances in using capsules at the same level to represent different properties appearing in different instances and using capsules at lower levels to compose higher-level capsules. The activation and composition processes of capsules are determined during inference, which is similar to previous structured models.
Motivated by this observation, we propose to combine the merits of both the traditional compositional model and the end-to-end trainable CNN via the capsule network~\cite{capsule} to obtain a model that possesses the compositional generation ability and performs well under the general settings. Although the capsule network~\cite{capsule, EMcapsule} demonstrates interesting grouping properties in some toy experiments, it still shows unsatisfactory results on the large-scale image datasets since the diverse visual compositions cannot be captured by learning grouping weights in a black-box manner without a proper guidance.

\begin{figure*}[t!]
	\centering
	\includegraphics[width=1.00\textwidth]{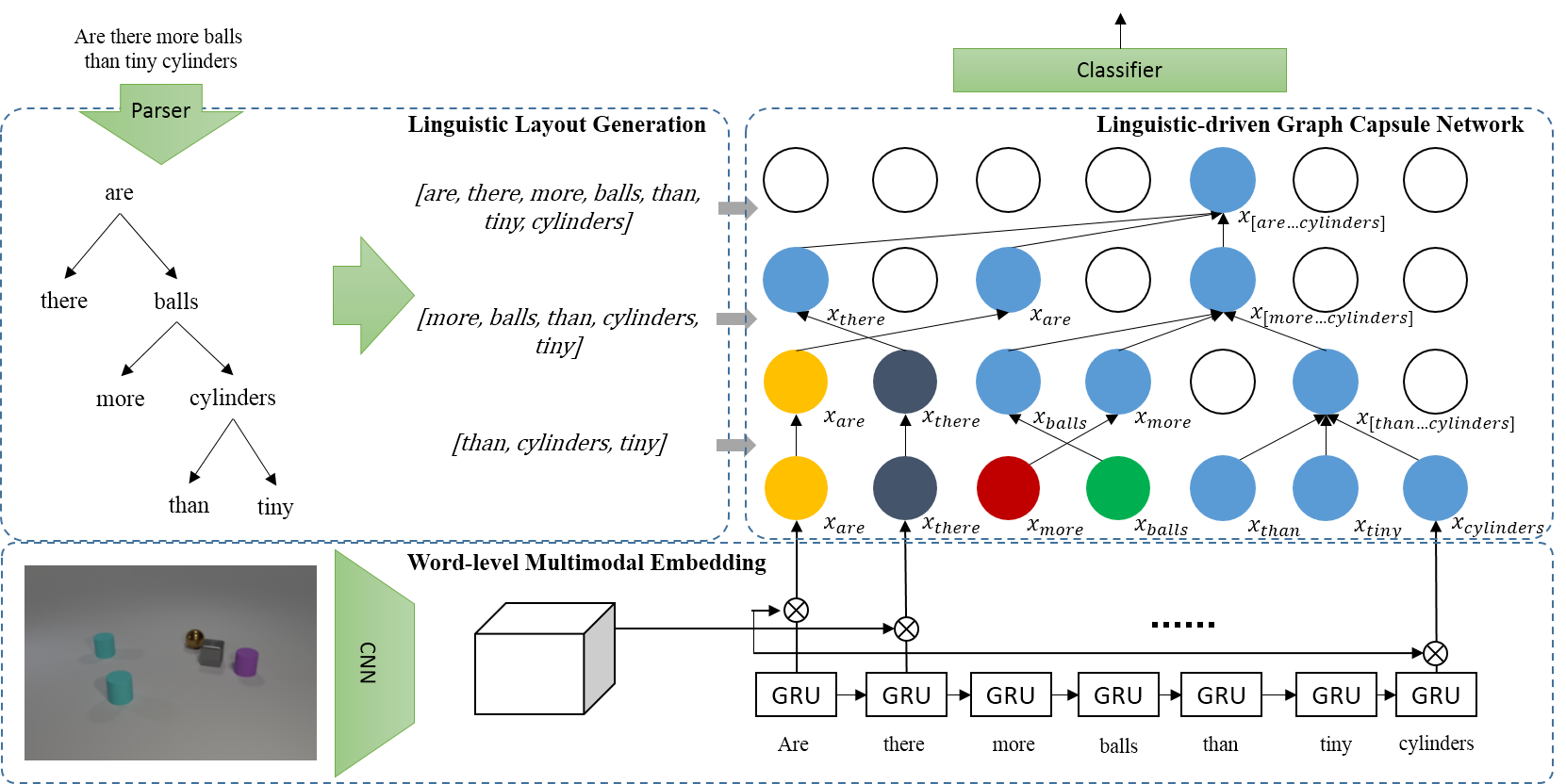}
	\caption{Pipeline of our model. We first parse the question into a tree and transform the tree into a linguistic layout. The nodes and their subtrees are grouped based on their distances from the furthest leaf nodes. The image feature is extracted from a pretrained CNN and fused with each word's encoding. The resulting feature maps can be the lowest layer in our model. In each layer, we forward the feature maps with the capsule layers and the corresponding linguistic guidance. We use color to indicate the merging groups. Capsules of the same color should be merged, and those of different colors should be routed to different capsules in the next layer. The hollow capsules indicate that they have not been activated. After forwarding of the capsule network from bottom to top, a parse tree-like structure, represented by the blue capsules, is carved.
}
	\label{fig:overview}
	\vspace{-5mm}
\end{figure*}

We thus propose a novel Linguistically driven Graph Capsule Network (LG-Capsule) to learn the adaptive reasoning routines for each image-question pair by guiding the capsule network via each individual parsed linguistic layout. Each linguistically driven capsule in our LG-Capsule network contains a multimodal representation of the visual evidence and a fragment of the question. Our proposed LG-Capsule network then performs transformation and merges routing for these capsules across multiple layers. The capsules in one layer are learned and merged from the bottom to the top, leading to fragments represented by capsules being combined from words into phrases, clauses and finally a sentence. In contrast to the original capsule network that clusters the lower-level capsules by agreement, our merging process is guided by the semantic parse tree of the question. For each capsule, our LG-Capsule network predicts a unary potential that indicates the probability of which higher-level capsule can be activated. Furthermore, for each pair, a binary potential is assigned to indicate whether the capsules should be routed together or not. The end-to-end conditional random field (CRF) then maximizes the probabilities and uses the final result as routing weights. 
As the random field is a fully connected graph with nodes representing the capsules and edges representing linguistic guidance, we name it linguistically driven graph routing. After forwarding all the capsule layers, the structured parse tree with soft connections is carved inside the CNN and entangled with visual patterns. The obtained complete question-image embedding is used to predict the final answer.

Our contribution can be summarized as follows. 1) We propose an end-to-end trainable LG-Capsule network that can incorporate external structured information into retain the good compositional generalization capability while maintaining performance on general tasks. 2) We propose utilizing a linguistic parse tree to guide the LG-Capsule network and tailor it to the visual question answering task. Extensive experiments have shown that our LG-Capsule network obtains comparable results on the CLEVR dataset and achieves a superior performance on the CLEVR compositional generalization test and the FigureQA dataset, which demonstrates its capability for both fully supervised and compositional generalization tasks.

\section{Related Works}
\textbf{Visual question answering.} The VQA task requires co-reasoning over both image and text to infer the correct answer.
Earlier works used the CNN-LSTM-based architecture and trained the neural networks in an end-to-end manner. Among the studies, those applying the attention mechanism~\cite{DBLP:journals/corr/IlievskiYF16, Xu2016,shih2016att, Visual7W, StackedAtt, HiCoAtt} have improved the answering accuracy by large margins on a natural image VQA benchmark~\cite{VQA}.
Subsequently, joint embedding of image and question representation~\cite{MCB, MLB, MFB, MUTAN} has been widely studied. The combination of the attention mechanism and compact bilinear multimodal fusion has further improved performance~\cite{Anderson2017up-down,vqav2winner}.
However, it has been argued that these black-box models might be attaining promising performance by exploiting the dataset bias instead of understanding the questions and images~\cite{balanced_vqa_v2, ECCV16baseline}. This argument has led to the proposal of unbiased datasets~\cite{balanced_vqa_v2,clevr,CountQA,IQA,DVQA,figureqa}; among them, CLEVR~\cite{clevr} has become a widely used benchmark of relational reasoning~\cite{RelNet,FiLM}.

Other strands of research have attempted to leverage information beyond image-question pairs, such as retrieving the external common knowledge and basic factual knowledge to answer the questions~\cite{externVQA, FVQA, Narasimhan_2018_ECCV}
or actively obtaining extra information and predicting the answer~\cite{zhu2017cvpr, Misra_2018_CVPR, IQA}. \cite{Li_2018_CVPR} tried to learn the complementary relationship between questions and answers and introduced question generation as a dual task to improve the VQA performance. ~\cite{Teney_2018_ECCV} dynamically selected example questions and answers during training, and encoded examples into a support set for answering the questions.

\textbf{Structured Reasoning on VQA.}
The recently proposed methods tried to incorporate structured information to address compositional visual reasoning while providing interpretation ability. Among them, neural modular networks~\cite{nmn,lnmn,e2emn} have attracted substantial interest. Rather than using a fixed structure, these studies use neural modules to solve a particular subtask and assemble them following a structured layout to predict the final answer.
Other approaches include structural attention, graph convolution networks and symbolic inference. For instance, ~\cite{SAVQA} was proposed to learn how to attend the image with CRF. 
\cite{norcliffebrown2018learning} applied a graph convolution network to obtain the question-specific graph representation and interactions in images. \cite{narasimhan2018out} also utilized a graph convolution network but embedded the retrieved knowledge with image representation as a node, and they evaluated their method on the FVQA~\cite{FVQA} knowledge-based VQA dataset. \cite{NeuralSymbolic} transformed images to scene graphs and questions to functional layouts and then performed symbolic inference on the graph. 

\textbf{Capsule network.}
Recently, Sabour~\emph{et~al.}~\cite{capsule} and Hinton~\cite{EMcapsule}~\emph{et~al.} proposed dividing each layer in a neural network into many small groups of neurons called ``capsules''.
The capsules can represent various properties of an object. specifically, a capsule can be activated if its represented property appears in a certain instantiation, and it can be routed to higher-level capsules iteratively to solve the problem. 
The intuition of local pattern activation and combination is reminiscent of previous structured models, such as the deformable part-based model~\cite{DPM}. In contrast to these models, the capsule network includes a neural network and has exhibited interesting properties on multiple datasets.

However, the existing capsule network studies only learn the grouping weights based on the discriminating loss only. They does not incorporate human priors and have not been evaluated on large-scale datasets. In this work, attempting to overcome this limitation, we leverage a human prior on language to guide the routing process and evaluate the capsule model on widely accepted VQA benchmarks.

\section{Linguistically driven Graph Capsule Network}

\subsection{Overview}
Given the free-form questions $Q$ and images $I$, our LG-Capsule network learns to predict the answers $y$.
As shown in Figure~\ref{fig:overview}, we first parse a question into a dependency parse tree using an off-the-shelf Stanford Parser~\cite{dep} and transform it to linguistic layout $G$. Then, we fuse the image feature and each word in the questions. Each resulting feature map is denoted by the capsule $x^0_i$, the encoded sentence fragments of which contains a single word of the question, and we denote it by sentence perception field $c^0_i$. Then, these capsules are concatenated as the input feature maps $X^0$ into our model.

Our model is constituted by several consecutive capsule layers. We denote the number of capsules by $n_c$; $x^l_i$ is the $i$-th capsule in layer $l$, and $c^l_i$ is its sentence perception field. The $l$-th layer receives inputs as the feature map $X^l=\{x^l_i\}_{i=1:n_c}$, the sentence perception field $C^l=\{c^l_i\}_{i=1:n_c}$ of capsules and the linguistic layout $g^{l+1}$ towards layer $l+1$. It performs routing and transformation on the input $X^l$ and the outputs feature maps $X^{l+1}$ and the corresponding perception field $C^{l+1}$.

Inside each layer, the routing process aims to select the several higher-level capsules that can best describe the lower-level capsules and decide which lower-level capsules should be merged together based on linguistic guidance.
Specifically, we use the unary potential $\psi(i)$ to indicate the probability of activating each higher-level capsule and the binary potential $\phi(i,j)$ to encourage $i$ and $j$ to select the same capsule if they are merged in the parsed tree. We maximize both potentials simultaneously by formulating a label assignment problem in a fully connected CRF, the nodes of which represent the capsules, and 
the nodes' labels represent the selection of higher-level capsules.
Thus, the routing weights are obtained by performing inference in a linguistically driven graph.

After forwarding through several layers, the tree-structured routing path is generated across these layers, and the last layer encodes the entire question-image embedding. We perform global average pooling on the last feature map and pass it through a multilayer perception to predict the final answer.


\subsection{Linguistic layout generation}\label{sec:layout}
We first generate the linguistic layout given the input question $Q$. We obtain the dependency parse tree by parsing the question with the off-the-shelf universal Stanford Parser~\cite{dep}. 
Next, we group the words according to whether they belong to the same subtree.

Specifically, we denote the node's level $l$ by the distance between that node and the furthest leaf node. 
Consider a subtree rooted at the node $i$ at the level $l$, and let $Tr(i)$ denote the set of words in this subtree; then, we group these words into a set and denote it by $g^l_i$. All sets that are at the same level form a list $g_l$. For example, we group nodes at the levels $0$ and $1$ into $g_0=\{\{are\},\{there\},\{more\},...,\{tiny\},\{cylinders\}\}$ and $g_1=\{\{than, tiny, cylinders\}\}$.
The generated layout $G=\{g_0,g_1,...,g_H\}$ is used to guide the aggregated routing process at different levels of capsule layers, where $H$ is the maximum level of the parse tree.

\subsection{Word-level multimodal embedding}
To obtain the input feature map for our model, we first extract the image feature $v$ and the word embedding vector $w$. The image feature $v$ is extracted from conv4 of ResNet-101~\cite{DBLP:conf/cvpr/HeZRS16} that was pretrained on ImageNet or a five-layer CNN trained from scratch.
The embedding vector $w_i$ for each word in the question is obtained via the gated recurrent network (GRU)~\cite{gru}. We first embed the word into a $200$-dimensional vector and then feed the entire question into the GRU. The final word embedding $\{w_i\}_{i=1:n_q}$ is the hidden vector of the GRU at its corresponding position, where $n_q$ is the number of words in the question $Q$. The hidden vector $w_{n_q}$ at the end of the question is also used as the question encoding $q$.

Next, we fuse the image feature $v$ with each word embedding vector $\{w_i\}_{i=1:n_q}$ using low-rank bilinear pooling~\cite{MLB}, which results in a sequence of multimodal representations $\{x_i^0\}_{i=1:n_q}$. Thus, each capsule $x_i^0$ in the layer $0$ is a multimodal representation of the image feature and a single word $c_i^0=\{w_i\}$. This feature map can be the input of our 
LG-Capsule network.

\begin{figure*}[t]
	\centering
	\includegraphics[width=1.00\textwidth]{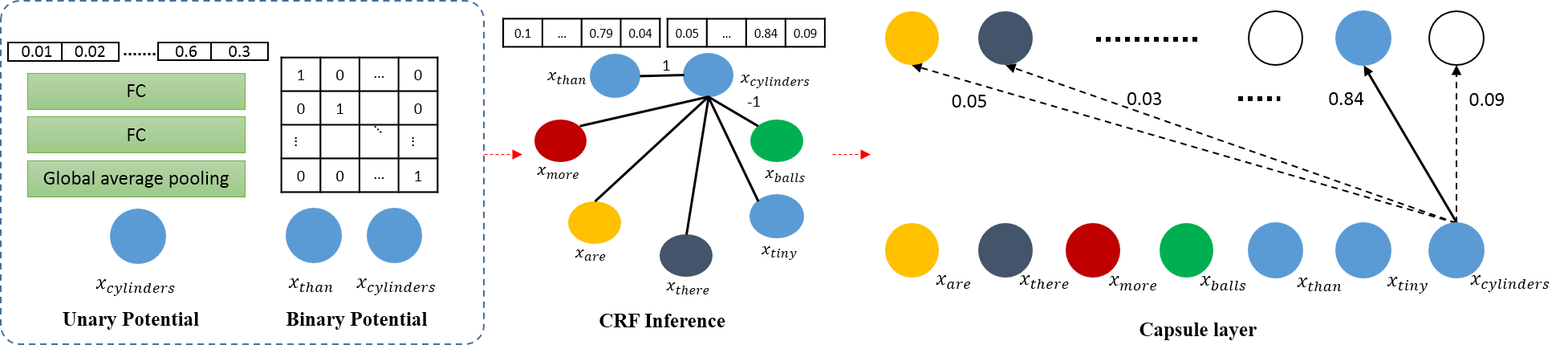}
	\caption{Routing process inside the capsule layer. We first generate the unary potential for each capsule and assign the binary potential for each pair of capsules based on the linguistic layout. Then, we build a fully connected graph and perform the CRF inference to maximize these two potentials across all capsules. The inference results are the routing weights of this layer. As illustrated, the words ``than'' and ``cylinders'' should be merged together; thus, the binary potential of their edge is $1$ if they select the same high-level capsule. The edges that do not connect to ``cylinders'' are omitted for clarity.
}
	\label{fig:routing}
	\vspace{-3mm}
\end{figure*}

\subsection{Linguistically driven tree routing} 
Given the input feature map $X^l$, the sentence perception field $C^l$ and the guided layout $g^{l+1}$, our capsule layer first generates the routing weight $r_{ij}^l$. Then, it generates the feature map $X^{l+1}$ and the perception field $C^{l+1}$ in the next layer.

The capsule layer generates $X^{l+1}$ by performing a linear transformation on the inputs $x_i^l$, which results in the vote $v_{ij}^l$ from the capsule $i$ to the capsule $j$ in the next layer $l+1$. The capsule layer also performs a nonlinear routing procedure $r_{ij}^l$ that indicates how much the capsule $i$ can influence $j$.
Formally,
\begin{equation}
\begin{aligned}
v_{ij}^l &= W_{ij} x_i^l \\
x_j^{l+1} &= \sum_i r_{ij}^l \ast v_{ij}^l
\end{aligned}
\label{eq:cap}
\end{equation}
where $x_j^{l+1}$ is the $j$-th capsule in the next layer.
Suppose that $r_{i}$ is a one-hot vector and $r_{ij}=1$; then, only the capsule $j$ can perceive the information from the capsule $i$. Further, if $r_{i'} = r_i$, then the capsule $j$ can merge the response from both the capsules $i$ and $i'$. 
Under this assumption, $c^{l+1}_j$ is the union of capsules' perception fields that have been routed to the capsule $j$, and $C^{l+1}$ can obtained with $c^{l+1}_j = \bigcup\limits_{j = \arg\max r_{i}^l} c^{l}_i$ for all $j$ in the layer $l$.
With several consecutive layers, it is natural to carve a tree-structured merging path inside a regular neural network.

The objective of the routing algorithm is to generate the weights $r_{ij}$, such that each lower-level capsule can activate a proper high-level capsule, and carve a structure isomorphic to the semantic parse tree. Formally, the perception field of capsules in each layer should be equivalent to the guided tree layout $C^{l+1} = g^{l+1}$.
We solve this problem by formulating the routing process as a label assignment problem in a CRF that consists of a unary term used to indicate the likelihood that the capsule $i$ chooses a specific capsule in the next layer independently and a binary term used to determine whether two capsules should select the same higher-level capsule, as shown in Figure~\ref{fig:routing}.

In the following, We first describe the details of our routing algorithm under the one-hot assumption. Then, we describe our method for generating routing weights and updating the word perception field softly.

\textbf{Unary potential}\label{sec:unary} \quad
The unary potential $\psi(i)$ indicates which higher-level capsule should be activated to represent the capsule $i$. We first perform the global max pooling on its feature map $x^l_i$ and then project it onto an $n_c$-dimensional vector with a neural network $f$ that consists of two fully connected layers. Here, $n_c$ is the number of capsules in the next layer. We apply softmax to normalize the resulting vectors such that each element is between $(0,1)$.

\textbf{Binary potential}\label{sec:binary} \quad
Given two capsules $i$ and $j$, they should be merged if there exists a parse tree node $a$ at the level $l$ that contains their merged sentence perception field: $c_{i}^l \cup c_{j}^l \supseteq g^{(l+1)}_a$.

The binary potential $\phi_{i,j}$ is an $n_c * n_c$ diagonal matrix used to enforce this structural constraint on capsule pairs. Potential $\phi_{i,j}(k_1,k_2)$ represents the potential of $i$ and $j$ having activated the high-level capsules $k_1$ and $k_2$, respectively.
\begin{equation}
\phi_{i,j}(k_1,k_2) = \begin{cases}
1     & k_1 = k_2 \land c_{i}^l \cup c_{j}^l \supseteq g^{(l+1)}_a  \\
-1    & k_1 = k_2 \land c_{i}^l \cup c_{j}^l \supseteq g^{(l+1)}_a  \\
0     & otherwise.
\end{cases}
\end{equation}
Intuitively, the binary potential $\phi(i,j)$ can be $1$ to encourage $i$ and $j$ to activate the same high-level capsule if they are merged in the parse tree or can be $-1$ to prevent them from merging early.

\begin{table*}[th!]
	\centering \small
	\begin{tabular}{|l|cc|ccc|c|}
		\hline
		Method & Count & Exist & \begin{tabular}[c]{@{}c@{}}Compare\\Integer\end{tabular} & \begin{tabular}[c]{@{}c@{}}Compare\\Attribute\end{tabular} & Query & Overall \\
		\hline
		N2NMN*~\cite{e2emn} & 68.5 & 85.7 & 84.9 & 88.7 & 90.0 & 83.7\\
		IEP*~\cite{inferring2017} & 92.7 & 97.1 & 98.7 & 98.9 & 98.1 & 96.9\\
		TbD+reg+hres*~\cite{Mascharka_2018_CVPR} & 97.6 & 99.2 & 99.4 & 99.6 & 99.5 & 99.1\\
		NS-VQA* (270 programs)~\cite{NeuralSymbolic} & 99.7 & 99.9 & 99.9 & 99.8 & 99.8 & 99.8 \\
		\hline
		CNN+LSTM+SAN~\cite{inferring2017} & 59.7 & 77.9 & 75.1 & 70.8 & 80.9 & 73.2\\
		LBP-SIG~\cite{SAVQA} & 61.3 & 79.6  & 80.7 & 76.3 & 88.6 & 78.0 \\
		Dependency Tree~\cite{Cao_2018_CVPR}  & 81.4 & 94.2 & 81.6 & 97.1 & 90.5 & 89.3\\
		CNN+LSTM+RN~\cite{RelNet} & 90.1 & 97.8 & 93.6 & 97.1 & 97.9 & 95.5\\
		CNN+GRU+FiLM~\cite{FiLM}  & 94.5 & 99.2 & 93.8 & 99.0 & 99.2 & 97.6\\
		MAC~\cite{hudson2018compositional} & 97.1 & 99.5 & 99.1 & 99.5 & 99.5 & 98.9\\
		\hline
		LG-Capsule (our method) & 95.6 & 98.7 & 97.2 & 98.8 & 98.8 & 97.9\\
		\hline
	\end{tabular}
	\caption{Comparison of question answering accuracy on the CLEVR dataset. The performance on question types \emph{Count}, \emph{Exist}, \emph{Compare Integer}, \emph{Compare Attribute} and \emph{Query} is reported in the respective columns. (*) indicates that the model has been trained with program annotations.
	}
	
	\label{table:clevr}
	\vspace{-3mm}
\end{table*}

\textbf{CRF inference}
The routing weights $r^l$ should maximize both the unary and binary potentials globally. This problem can be formulated as a labeling problem in a CRF, the nodes of which are the capsules in the current layer, and labels are the capsules in the next layer. An optimized labeling gives the corresponding route weights $C^l$. To enable training in an end-to-end manner, we use the method proposed in ~\cite{SAVQA} that implements the mean-field algorithm with a neural network.

\textbf{Soft perception field}
Generally, the routing weights $r_{i}$ cannot be a one-hot vector, and the maximum operation cannot be differentiated with respect to the index, which thus prevents the entire model from being trained end-to-end.
To address this problem, we relax the capsule perception field $c^l_i$. Instead of using $0$ or $1$ to represent which words have been encoded in the capsule $i$, we allow the field to perceive a word with a weight between $(0,1)$.

For example, suppose that $n_q$ is the length of the question; then, $c^l_i$ is a $n_q$-dimensional vector, the elements of which represent to what degree it encodes a word. 
At the input layer $0$, the capsule $i$ is the fusion between the image and $i$-th word in the question; then, $c^0_i$ is a one-hot vector with elements, all of which are $0$ except for the $i$-th entry. Then, if the softmax-normalized routing weights are $r_{ij}^0=0.9$ and $r_{ik}^0=0.05$, the sentence perception scores of the $i$-th word for the capsules $j$ and $k$ are $c^1_j = 0.9$ and $c^1_k = 0.05$, respectively.

Formally, we first transform the guided layout $G$ to an $n_q * n_q$ correlation matrix $G'$, representing the compatibility of two words at different layers, which is $1$ if they should be merged at layer $l$.
\begin{equation}
{g'}^l(i,j) = \begin{cases}
1     & \exists a \quad i,j \in g'^l_a  \\
-1    & \exists a,b \quad i \in g'^l_a,j \in g'^l_b, a\neq b  \\
0     & otherwise,
\end{cases}
\end{equation}
where $g$ is the linguistic layout described in Sec.~\ref{sec:layout}

Next, we need to obtain the binary potential. We use $\phi(i,j)$ to denote the compatibility of the capsules $i$ and $j$. The greater the number of compatible words contained by $i$ and $j$ is, the higher this value should be: $\phi(i,j) = {c^l_{j}}^T * {g'}^{l+1} * {c^l_{i}}$.
For all capsule pairs, we write the binary potential compactly in the matrix form:
\begin{equation}
\phi = {C^l}^T{g'}^{l+1}C^l
\end{equation} 
Then, we expand each binary potential $\phi(i,j)$ into a diagonal $n_c * n_c$ matrix as described above.
We use this binary potential and the unary potential to build the CRF, and obtain the routing weights for all the capsules $R^l$. Given $R^l$, we propagate the perception score conditioned on the routed weights:
\begin{equation}
C^{l+1} = {R^{l}}^TC^l
\end{equation}

By relaxing the sentence perception field to a perception score, we allow all structured components to be soft, and our model can be trained end-to-end while only using the answer as supervision.

\section{Experiment}
In this section, we validate the effectiveness and generalization capability of our models on the CLEVR dataset, the CLEVR computational generalization test, and the FigureQA dataset.

\subsection{Datasets}

\textbf{CLEVR}~\cite{clevr} is a synthesized dataset designed to achieve minimal biases and test compositional reasoning. It consists of $100,000$ images, $853,554$ questions and the corresponding image scene graphs and questions' functional program layouts. Given the scene graphs, the images are rendered using objects of random shapes, colors, materials and sizes. The questions are generated based on functional program layouts that consist of functions that can select a certain color, a shape, or compare two objects. 
Thus, a VQA model should be able to encode the targeted objects and their relations to answer the questions correctly.

The \textbf{CLEVR composition generalization test (CLEVR-CoGenT)}~\cite{clevr} is proposed to investigate the composition generalization ability of a VQA model. This dataset contains synthesized images and questions similar to those in CLEVR but has two conditions: in condition A, all cubes are gray, blue, brown, or yellow, and all cylinders are red, green, purple, or cyan. In condition B, cubes and cylinders swap color palettes. Thus, a model cannot achieve good performance on condition B by simply memorizing and overfitting the samples in condition A.

\textbf{FigureQA}~\cite{figureqa} is also a synthesized dataset. This dataset contains $100,000$ images and $1,327,368$ questions for training. In contrast to CLEVR, the images are scientific-style figures. The dataset includes five classes: line plots, dotted-line plots, vertical and horizontal bar graphs, and pie charts. The questions also concern various relationships between elements in the figures, such as the maximum and the area under the curve. Similar to CLEVR-CoGenT, this dataset also has two conditions. A total of $100$ unique colors are divided into two sets; one is used on vertical bar graphs, line charts, and pie charts, and the other is used on horizontal bar graphs and dotted-line charts. The color palettes are swapped between two conditions to evaluate models on unseen color combinations.

\subsection{Implementation details}\label{sec:details}
For the CLEVR and CLEVR-CoGenT datasets, we employ the same settings as those used in \cite{inferring2017,clevr} to extract the image feature and word encoding. The images are resized to $224 \times 224$, and $1024 \times 14 \times 14$ feature maps are extracted from conv4 of ResNet-101 pretrained on ImageNet. The 
$1024$-dimensional feature maps are concatenated with a $2$-channel coordinate map and are projected onto a $128$-dimensional space using a single $3 \times 3$ convolutional layer. 
The words are first embedded as $200$-dimensional vectors. Then, we apply a bidirectional GRU with $512$-dimensional hidden states for both directions to extract the words' encoding vector.
The $128 \times 14 \times 14$ feature maps are fused with the $1024$-dimensional word embedding using bilinear pooling~\cite{MLB}.
The questions in CLEVR and CLEVR-CoGenT datasets have the maximum length of $46$, and the heights of their parse trees are mostly less than $4$. Thus, we prune the parse trees, keep the top-$4$ levels, and set the number of capsule layers to $4$. The input feature maps at the layer $0$ have a total of $46$ capsules, and each capsule is a $128 \times 14 \times 14$ feature map. Because the maximum number of the level-$1$ nodes in the parsed tree is $9$, we set the number of capsules produced by each layer to $9$; each capsule has $16$ feature channels. 
Lastly, the classifier can convolve the $144$-dimensional feature maps to $512$ dimensions and feed the result into two fully connected layers with output sizes of $1024$ and $29$, where $29$ is the number of candidate answers. 

We resize the images of FigureQA to $256 \times 256$. Then, we use a five-layer CNN to extract the image features. Each layer has a $3 \times 3$ kernel and a stride of $2$. The dimensions of the feature map in the first four layers are $64$, and the last layer has $128$-channel outputs. 
Words are embedded into $1024$-dimensional vectors, and $128 \times 8 \times 8$ feature maps are fused with the $1024$-dimensional word embedding, which is the same as in the CLEVR dataset.
The questions in FigureQA have the maximum length of $11$, and the maximum heights of their parse trees are $6$. Thus, in FigureQA, our model has $6$ capsule layers, and layer-$0$ feature maps have $11$ capsules. We set the number of capsules of the following feature maps to $8$; each has $16$-dimensional features.
Lastly, the classifier projects $128$-dimensional feature maps to $512$, $1024$ and $2$ dimensions to output the final answers.

\begin{table}[t] \centering
	\resizebox{\columnwidth}{!}{%
		\begin{tabular}{|l|c|c|}
			\hline
			& \multicolumn{2}{c|}{Training on A} \\    
			Model & \multicolumn{1}{c}{Testing on A} & \multicolumn{1}{c|}{Testing on B}  \\
			\hline
			IEP~\cite{inferring2017} & 96.6 & 73.7  \\
			NS-VQA~\cite{NeuralSymbolic} & 99.8 & 63.9 \\
			NS-VQA+Ori~\cite{NeuralSymbolic} & 99.8 & 99.7\\
			\hline
			CNN+LSTM+SA~\cite{inferring2017} & 80.3 & 68.7  \\
			CNN+GRU+FiLM~\cite{FiLM} & 98.3 & 75.6  \\
			CNN+GRU+FiLM 0-Shot~\cite{FiLM} & 98.3 & 78.8  \\
			TbD+reg~\cite{Mascharka_2018_CVPR} & 98.8 & 75.4 \\
			LG-Capsule & 98.34 & 84.06 \\
			\hline
		\end{tabular}
	}
	\caption{Comparison of question answering accuracy on the CLEVR-CoGenT validation set. Each method is trained on condition A only, and is evaluated on both conditions A and B.}
	\label{table:clevr_CoGenT}
	\vspace{-5mm}
\end{table}

To reduce the computational complexity for CLEVR and FigureQA, we prune the leaf nodes that are not nouns in the parse tree and prune the leaf nodes that are neither nouns nor words denoting colors for CLEVR-CoGenT.
For all three of these datasets, we align the questions to the right by padding with $0$s from the left and align the linguistic layout to the bottom. We train our model on the training set and evaluate it on the validation and test sets. 
The model is trained with the Adam optimizer~\cite{adam}. The base learning rate is $0.0003$ for CLEVR and CLEVR-CoGenT and is $0.0001$ for FigureQA. The batch size is $64$. The weight decay parameters ${\beta}_1$ and ${\beta}_2$ are $0.00001$, $0.9$, and $0.999$ , respectively.

\begin{table}[t] \centering
	\center
	\begin{tabular}{|l|*{2}{c|}}
		\hline
		Model & Val. & Test \\
		\hline
		Text only~\cite{figureqa} & 50.01 & 50.01\\
		CNN+LSTM~\cite{figureqa} & 56.16 & 56.00\\
		CNN+LSTM on VGG-16 features~\cite{figureqa} & 52.31 & 52.47\\
		RN~\cite{figureqa} & 72.54 & 72.40\\
		LG-Capsule & 90.58 & 90.70 \\ 
		\hline
	\end{tabular}
	\caption{Comparison of question answering accuracy on the FigureQA validation and test sets that have alternative color schemes.}
	\label{table:figureqa}
	\vspace{-5mm}
\end{table}

\begin{figure*}[t!]
	\centering
	\includegraphics[width=1.00\textwidth]{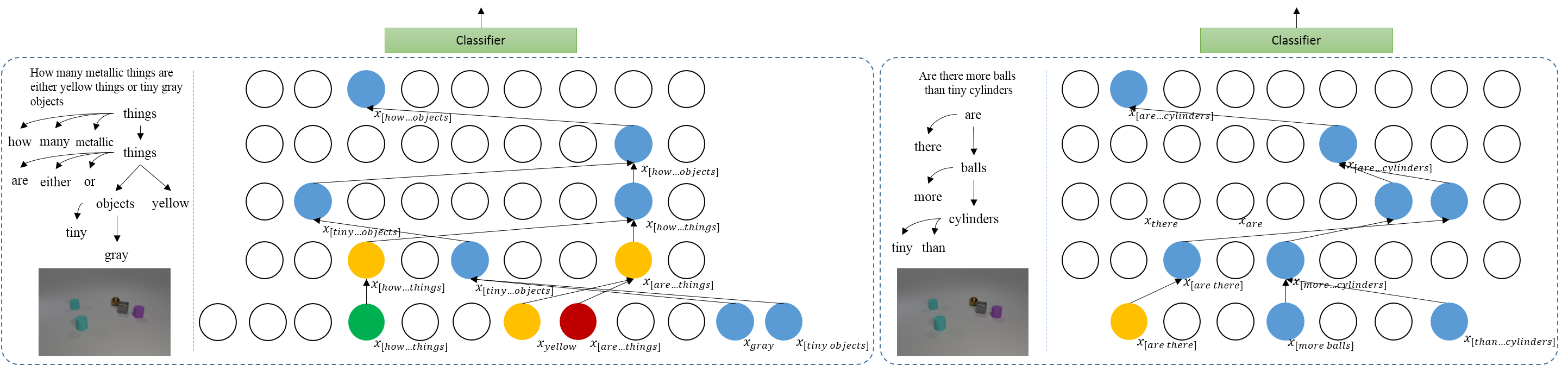}
	\caption{Two visualization examples of our routing results on the CLEVR-CoGenT condition B validation set. The questions, parse tree and the image are shown on the left, and the routings in the LG-Capsule network are shown on the right. We display all $9$ capsules per layer and all $4$ capsule layers as described in Section~\ref{sec:details}. The curved edge in the parse tree indicates the leaf nodes that have been pruned to reduce complexity. We omit the left-padding capsules. Thus, the number of capsules displayed at layer $0$ is equal to the question length, and the capsules corresponding to the pruned leaf nodes are hollow. The same color of capsules indicates that they should be merged, and different colors indicate the opposite. The arrows between capsule layers show the selection of a high-level capsule that has the largest routing weights, given a specific lower-level capsule.}
	\label{fig:vis}
	\vspace{-2mm}
\end{figure*}

\subsection{Comparison with the state-of-the-art methods}
\subsubsection{CLEVR dataset}
Table~\ref{table:clevr} shows the performance of all the compared methods on the CLEVR test set. The end-to-end modular network~\cite{e2emn}, program execution engine~\cite{inferring2017}, transparency by design~\cite{Mascharka_2018_CVPR}, and neural symbolic visual question answering~\cite{NeuralSymbolic} are referred to as ``N2NMN'', ``IEP'', ``TbD'' and ``NS-VQA'', respectively. 
All these methods use the functional programs' layout as the extra training signal. 
``N2NMN'', ``IEP'' and ``TbD'' achieve their best results by using all of the program layouts. Although ``NS-VQA'' uses only $3$ samples from each of the $90$ question families, it leverages the scene graphs to train a scene parser. ``N2NMN', ``IEP'' and ``NS-VQA'' also have variants that use different numbers of programs during training, and it has been shown that their performance degrade if fewer program layout examples are used.

In contrast, our model can achieve a comparable state-of-the-art performance without using any dataset-specific layout, thus illustrating the visual question answering ability of our LG-Capsule network in the regular setting. We further demonstrate the advantage of our model in compositional generalization ability in the following experiments.

\subsubsection{CLEVR compositional generalization test}
We report the answering accuracy values of different models on CLEVR-CoGenT in Table~\ref{table:clevr_CoGenT}. The accuracy values are obtained by training the models in Condition A, and evaluating accuracy in both Condition A and Condition B. As shown in Table~\ref{table:clevr_CoGenT}, while achieving a comparable accuracy in Condition A our proposed LG-Capsule network significantly ,outperforms all the compared methods except NS-VQA+Ori~\cite{NeuralSymbolic} by large margins in Condition B without fine-tuning on the alternative color scheme. Note that, NS-VQA+Ori requires to use both scene graph and a question's functional layout as additional supervised signals. Without additional training signals as our model for a fair comparison, its accuracy downgrades into 63.9\%, i.e., nearly 20\% worse than our model (63.9 vs 84.06). This verifies the superiority of our model over all the compared methods in terms of the composition generalization ability.

\subsubsection{FigureQA dataset}
Table~\ref{table:figureqa} presents the comparison of our model and the existing methods on the FigureQA dataset. The baseline methods ``Text only'', ``CNN+LSTM'', ``CNN+LSTM on VGG-16 features'' and ``RN~\cite{RelNet}'' are originally from~\cite{figureqa}. As shown, our LG-Capsule network surpasses the best of all compared methods (i.e., the relational network that performs very well on the CLEVR dataset) by a large margin ($18.3\%$). Considering the property of FigureQA and its difference to the CLEVER dataset, this result demonstrates that our model can consistently performs well across different datasets and verifies the superior generalization ability of our model over all the compared methods.

\begin{table}[t] \centering
	\center
	\resizebox{\columnwidth}{!}{%
		\begin{tabular}{|l|c|c|c|c|}
			\hline
			& \multicolumn{2}{c|}{CLEVR-CoGenT} & \multicolumn{2}{c|}{FigureQA} \\
			
			Model & \multicolumn{1}{c}{A} & \multicolumn{1}{c|}{B} & \multicolumn{1}{c}{A} & \multicolumn{1}{c|}{B}  \\
			\hline
			Baseline (CNN) & 97.59 & 78.19 & 88.73 & 87.03 \\
			+unary & 97.21 & 83.22 & 90.34 & 89.09 \\
			+unary+binary & 93.61 & 78.29 & 87.53 & 85.09 \\
			+unary+binary+Parser~(LG-Capsule) & 98.34 & 84.06 & 92.07 & 90.58 \\
			\hline
		\end{tabular}
	}
	\caption{Answering accuracy values of variants of our method on CLEVR-CoGenT and FigureQA validation sets. We incrementally change the routing process for the baseline CNN model, and report the results in respective rows.}
	\vspace{-5mm}
	\label{table:ab}
\end{table}

\subsection{Ablation Studies}
We evaluate the effectiveness of our linguistically driven routing process on CLEVR-CoGenT and FigureQA. The results are shown in Table.~\ref{table:ab}.
The baseline model is a regular convolutional neural network without routing. This CNN has exactly the same architecture as that of our model. The row ``+unary'' shows the results of our model that uses only the unary potential for routing. The capsules in this model variant select the high-level capsules independently and thus do not require CRF inference. The model variant ``+unary+binary'' learns the binary potential based on the answer classification loss only. We concatenate the features of each capsule's pairs and use two fully connected layers to predict the $n_c * n_c$ matrix described in section~\ref{sec:binary}.
The last row ``+unary+binary+Parser'' represents our LG-Capsule model that uses the parsed tree to generate the binary potential.
Table~\ref{table:ab} shows that different model variants achieve the similar accuracy on Condition A; however, the accuracy on unseen examples in Condition B varies considerably. Our model outperforms the baseline by $5.87\%$ and $3.55\%$ on CLEVR-CoGenT and FigureQA. These performance gains demonstrate the superiority of our linguistically driven routing process in terms of compositional generalization ability.


\subsection{Visualization of routing results}
We visualize our routing results in Figure~\ref{fig:vis}. The input questions, image, and linguistic guidance are shown on the left, while the routing results are shown on the right.
The first example combines terms ``gray'' and ``objects'', matching the parse tree. However, it first combines ``yellow objects'' with ``how many'', encodes ``yellow objects'' and ``gray objects'' separately, and then combines them to predict the answer at last. The example follows the linguistic guidance at first but demonstrates a more reasonable routing process than the parse tree to answer the question.
The second example merges the terms ``cylinders'', ``balls'' and ``are'' and carves a tree similar to the parse tree.

\section{Conclusion}
We propose a novel Linguistically driven Graph Capsule Network (LG-Capsule) that can be trained in an end-to-end manner while incorporating linguistic information to achieve compositional generalization capability. Specifically, we use the unary potential for each capsule to activate proper high-level capsules and the binary potential for each capsule pair to incorporate a linguistic prior into the questions' structures. The end-to-end CRF is applied to maximize two types of potential, and the final results are used as routing weights.
As we bind the lowest capsule with a single word and a visual feature, the bottom-up linguistically guided merging process can lead from combining words to phrases, clauses and finally a sentence. After forwarding all capsule layers, the parse tree with soft connections is carved inside the CNN and entangled with visual patterns. In the future, we will progressively refine our model to further enrich its generalization performance and discriminative power.


\ifCLASSOPTIONcaptionsoff
  \newpage
\fi


\bibliographystyle{IEEEtran}
\bibliography{egbib}

\begin{IEEEbiography}[{\includegraphics[width=1in,height=1.25in,clip,keepaspectratio]{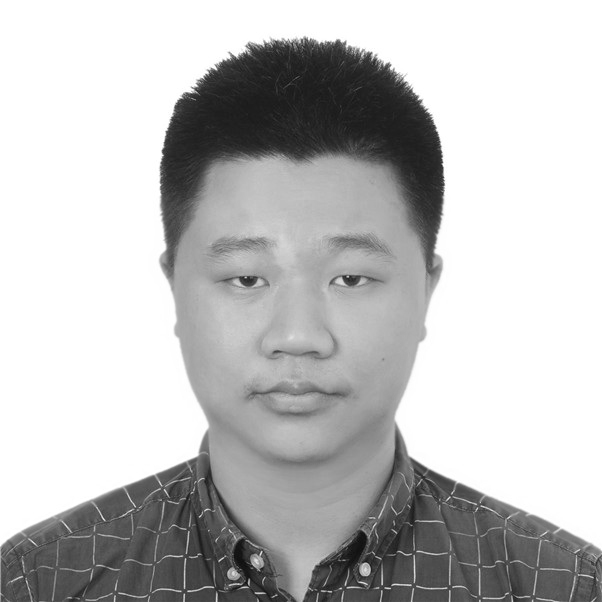}}]{Qingxing Cao} is currently a postdoctoral researcher in the School of Intelligent Systems Engineering at Sun Yat-sen University, working with Prof. Xiaodan Liang. He received his Ph.D. degree from Sun Yat-Sen University in 2019, advised by Prof. Liang Lin. His current research interests include computer vision and visual question answering.
\end{IEEEbiography}
\begin{IEEEbiography}[{\includegraphics[width=1in,height=1.25in,clip,keepaspectratio]{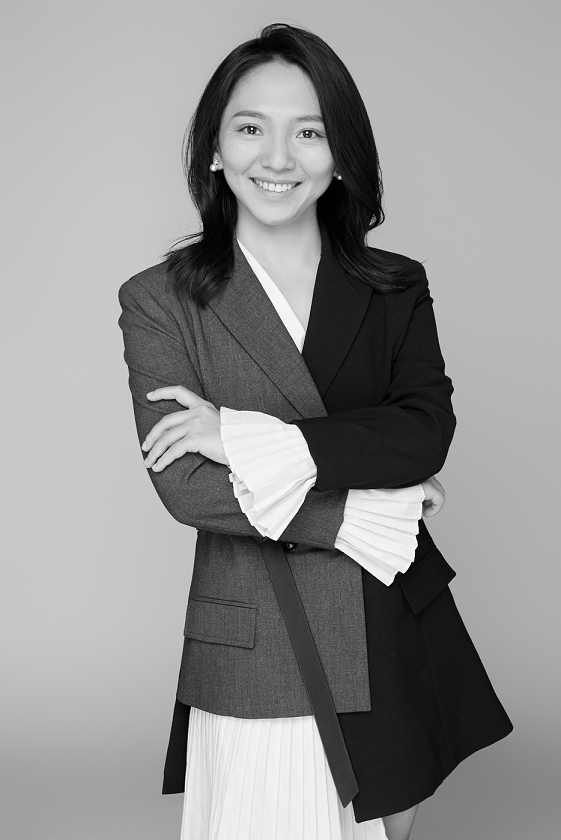}}]{Xiaodan Liang} is currently an Associate Professor at Sun Yat-sen University. She was a postdoc researcher in the machine learning department at Carnegie Mellon University, working with Prof. Eric Xing, from 2016 to 2018. She received her PhD degree from Sun Yat-sen University in 2016, advised by Liang Lin. She has published several cutting-edge projects on human-related analysis, including human parsing, pedestrian detection and instance segmentation, 2D/3D human pose estimation and activity recognition.
\end{IEEEbiography}

\begin{IEEEbiography}[{\includegraphics[width=1in,height=1.25in,clip,keepaspectratio]{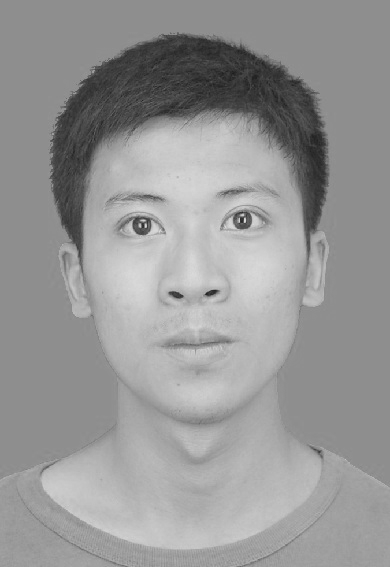}}]{Keze Wang} received his B.S. degree in software engineering from Sun Yat-Sen University, Guangzhou, China, in 2012. He is currently pursuing his dual Ph.D. degree at Sun Yat-Sen University and Hong Kong Polytechnic University, advised by Prof. Liang Lin and Lei Zhang. His current research interests include computer vision and machine learning.
\end{IEEEbiography}

\begin{IEEEbiography}[{\includegraphics[width=1in,height=1.25in,clip,keepaspectratio]{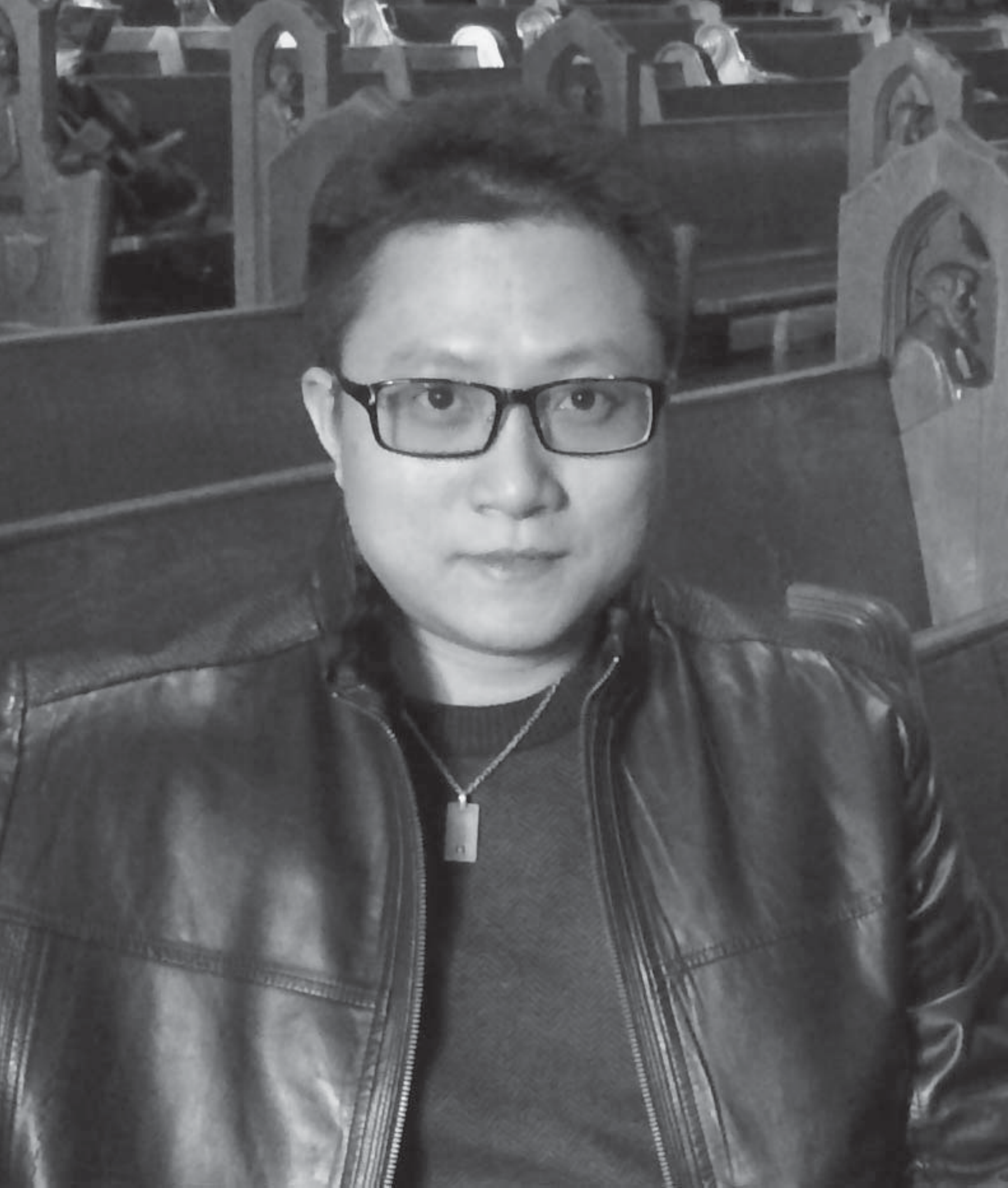}}]{Liang Lin} is a full professor of Computer Science in Sun Yat-sen University. He served as the Executive Director of the SenseTime Group from 2016 to 2018, leading the R\&D teams in developing cutting-edge, deliverable solutions in computer vision, data analysis and mining, and intelligent robotic systems.  He has authored or co-authored more than 200 papers in leading academic journals and conferences (e.g., TPAMI/IJCV, CVPR/ICCV/NIPS/ICML/AAAI). He is an associate editor of IEEE Trans, Human-Machine Systems and IET Computer Vision, and he served as the area/session chair for numerous conferences, such as CVPR, ICME, ICCV, ICMR. He was the recipient of Annual Best Paper Award by Pattern Recognition (Elsevier) in 2018, Dimond Award for best paper in IEEE ICME in 2017, ACM NPAR Best Paper Runners-Up Award in 2010, Google Faculty Award in 2012, award for the best student paper in IEEE ICME in 2014, and Hong Kong Scholars Award in 2014. He is a Fellow of IET.
\end{IEEEbiography}
\end{document}